\documentclass{article}

\usepackage{microtype}
\usepackage{graphicx}
\usepackage{subcaption}
\usepackage{booktabs} 
\usepackage{enumitem}
\usepackage{tabularx}
\usepackage{titlesec}

\titlespacing*{\paragraph}{0pt}{2pt}{6pt}

\usepackage{hyperref}

\usepackage[accepted]{icml2026}

\usepackage{amsmath}
\usepackage{amssymb}
\usepackage{mathtools}
\usepackage{amsthm}

\usepackage[capitalize,noabbrev]{cleveref}

\theoremstyle{plain}

\theoremstyle{definition}

\theoremstyle{remark}

\usepackage[textsize=tiny]{todonotes}

\icmltitlerunning{ISM: Self-Improving Strategy Memory for Continual Mathematical Reasoning}

\begin{document}

\twocolumn[
  \icmltitle{ISM: Self-Improving Strategy Memory for Continual Mathematical Reasoning}



\begin{icmlauthorlist} \icmlauthor{Prakhar Dixit}{umbc} \icmlauthor{Tim Oates}{umbc} \end{icmlauthorlist} \icmlaffiliation{umbc}{ Department of Computer Science, University of Maryland, Baltimore County, Baltimore, Maryland, USA } \icmlcorrespondingauthor{Prakhar Dixit}{pdixit1@umbc.edu} \icmlcorrespondingauthor{Tim Oates}{oates@umbc.edu} \icmlkeywords{ Large Language Models, Continual Learning, Mathematical Reasoning, External Memory } \vskip 0.3in ] \printAffiliationsAndNotice{}
\begin{abstract}
We propose \textbf{Intelligent Schema Memory (ISM)}, a self-evolving memory-augmented system that improves mathematical reasoning for a frozen LLM under continual learning with hard episodic resets. ISM maintains a compact, self-refined bank of strategy schemas learned from both successful and failed episodes, with symbolic tools that check intermediate steps and certify answers. Without updating model parameters, ISM outperforms passive, retrieval, and reflection baselines on MATH-Hard and OlympiadBench, using 64\% and 86\% fewer schemas respectively than the strongest passive baseline. These results show that small, actively maintained, and verified strategy memories can support reliable continual mathematical reasoning under strict episodic isolation. The codebase is available at \url{https://github.com/pdx97/ISM}.
\end{abstract}

\begin{figure*}[!t]
    \centering
    \includegraphics[width=0.9\linewidth]{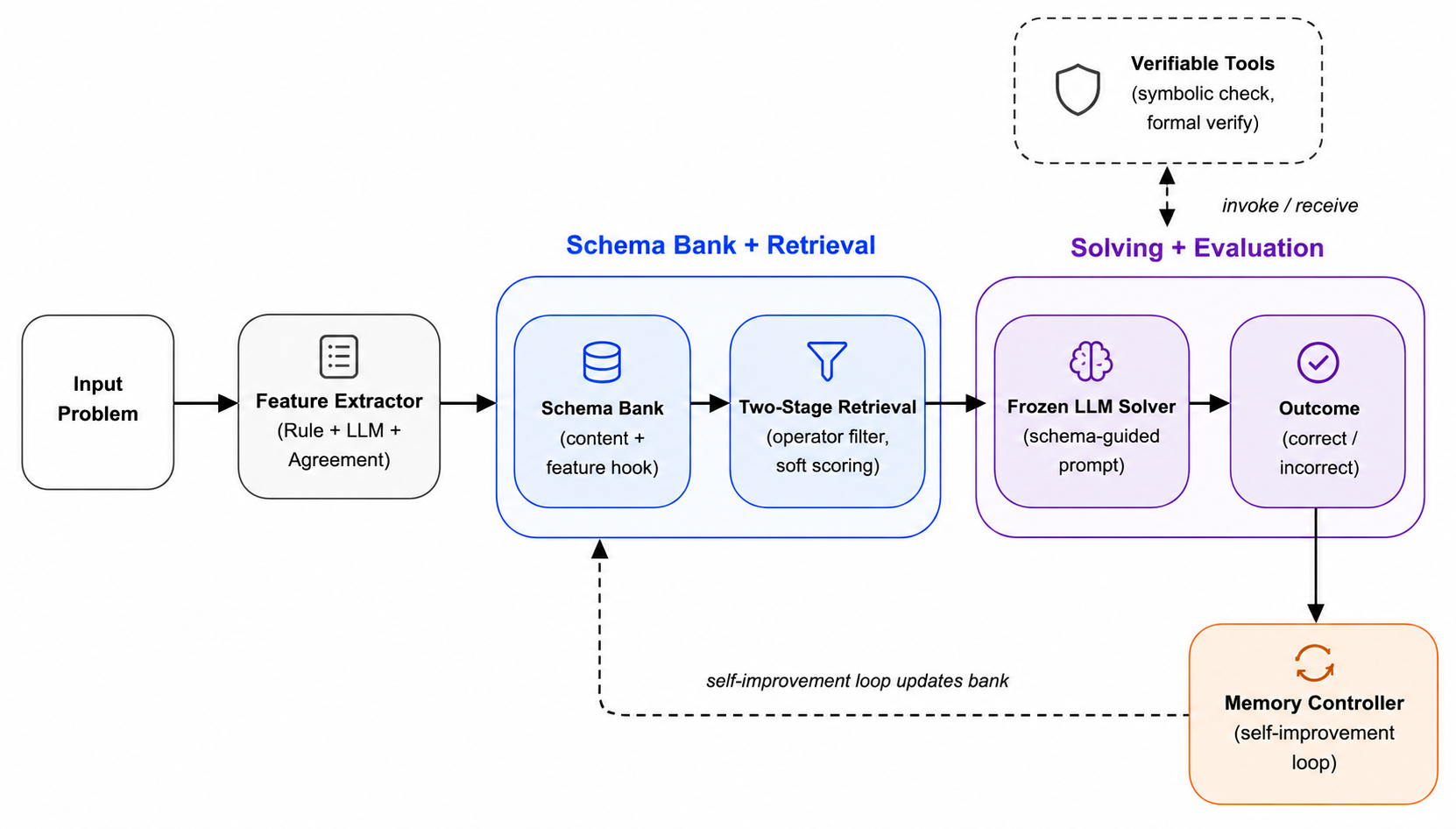}
    \caption{Pipeline of the ISM system. A problem passes through the feature extractor (rule-based and LLM-based with agreement scoring), then the schema bank with two-stage retrieval (operator filter and soft scoring). The retrieved schema's content is injected into the frozen LLM solver, which may invoke verifiable tools. After each episode, the Memory Controller's self-improvement loop (dashed arrow) updates the schema bank.}
    \label{fig:pipeline}
\end{figure*}

\section{Introduction}

Large language models (LLMs) \cite{zhao2026surveylargelanguagemodels} excel at solving isolated mathematical problems, but they frequently stumble when faced with a sequential stream of problems spanning shifting domains. Because each new problem starts from the same frozen parameters, any insights the model gained from a previous task are lost before the next one begins. This highlights the core challenge of continual learning under a frozen backbone: any cumulative improvement must exist entirely outside the model, within an external repository of knowledge. Mathematical reasoning provides an ideal testbed for studying this dynamic, given its clear structural rules and verifiable correctness. Moreover, engineering a system that improves over time without retraining addresses a fundamental question in AI: \emph{How can a frozen LLM self-evolve at mathematical reasoning while keeping its accumulated knowledge bounded, verifiable, and reliable?}

Existing approaches address aspects of this challenge but leave critical gaps. Reflection-based methods like Reflexion \cite{shinn2023reflexion} generate verbal feedback following a failure, but this feedback remains free text rather than a reusable, structured strategy. Retrieval-augmented setups simply store raw past examples; they grow without bound and rely entirely on surface-level embedding similarity to determine what to retrieve. Meanwhile, skill-library systems like Voyager \cite{wang2023voyager} build executable repositories of learned behaviors, but these libraries only expand. They lack mechanisms to verify new entries, consolidate near-duplicates, or retire obsolete strategies. Consequently, current systems fail to self-improve in a controlled manner; they either forget past lessons, bloat their storage, or accumulate noise.

To address these limitations, we propose \textbf{Intelligent Schema Memory (ISM)}, a self-evolving memory framework that retrieves strategies and actively regulates its repository via an autonomous self-improvement loop, all while keeping the underlying LLM frozen. ISM maintains a compact bank of strategy schemas. Each schema is split into two components: a \textit{content section} containing the strategy description, solution template, and heuristics injected into the solver prompt; and a \textit{feature hook} containing structural and embedding-based retrieval signals that adapt online based on usage. This separation decouples what a schema knows from how it is discovered, allowing retrieval accuracy to sharpen with experience without altering the underlying strategies.

The memory bank is maintained by seven self-improvement mechanisms: four govern quality and footprint (audit, merge, prune, and promote/demote), two learn symmetrically from outcomes (\textit{reinforce} strengthens successful paths, while \textit{antipattern} records failures to avoid), and one rehabilitates underperforming schemas before they face pruning. Crucially, every memory update is gated by symbolic verification, preventing the bank from reinforcing faulty generalizations. Together, these mechanisms transform external memory from a passive storage unit into an actively maintained, bounded, and self-correcting asset. The combination of symmetric success/failure learning, online-adapting retrieval, lifecycle management, and verification gating is what distinguishes ISM from prior memory-augmented systems built around frozen LLMs.

We evaluate ISM over a 300-episode continual learning stream \cite{wang2024comprehensive} on two competition-level benchmarks, MATH-Hard \cite{hendrycks2021math} and OlympiadBench \cite{he2024olympiadbench}, against five baselines: vanilla, retrieval, reflection, static schema, and passive schema memory. ISM achieves the highest accuracy on both benchmarks while storing 64\% and 86\% fewer schemas than the strongest passive baseline, and up to 23$\times$ fewer than retrieval-based systems. This performance gap widens as the stream progresses: ISM exhibits greater resilience to domain shifts than any baseline, and its memory bank remains strictly bounded while retrieval-based storage grows linearly with the number of problems encountered.\textbf{Our main contributions are}:
\begin{itemize}
    \item We introduce \textbf{ISM}, a self-evolving memory architecture with an autonomous self-improvement loop that retrieves strategies, learns symmetrically from successes and failures, and gates every update through symbolic verification.
    \item We propose a dual-representation schema design that decouples content (what to do) from retrieval features (when to apply it), letting retrieval adapt online while strategy content stays stable.
    \item We evaluate ISM under a continual learning protocol with hard episodic resets on MATH-Hard and OlympiadBench, where it outperforms five baselines while using up to 23$\times$ less storage.
\end{itemize}

\section{Related Work}

\paragraph{In-model memory and continual learning.}
A recent line of work addresses continual learning by redesigning the model's internal memory substrate. Titans~\cite{behrouz2024titans} introduces a neural long-term memory module that learns to memorize at test time, partitioning representation between a short-term attention buffer and a long-term recurrent memory whose parameters update during inference. Nested Learning~\cite{behrouz2025nested} generalizes this perspective, recasting a model as a hierarchy of nested optimization problems with distinct update frequencies, fast inner loops for transient context and slow outer loops for stable knowledge, instantiated in the HOPE architecture and its Continuum Memory System. These approaches mitigate catastrophic forgetting by enriching the model's own memory substrate at the architecture level. ISM is complementary: rather than redesign the substrate, we keep the LLM entirely frozen and externalize the memory hierarchy as a symbolic schema bank with explicit lifecycle management. The two directions are largely orthogonal and could in principle be combined.
\paragraph{External memory and skill libraries.}
Several systems attach external memory to frozen models. MemGPT~\cite{packer2024memgpt} introduces tiered context windows backed by a searchable archive, and agentic memory systems organize past episodes through dynamic structures. Skill-library systems such as Voyager~\cite{wang2023voyager} accumulate executable skills as an agent operates in an environment, appending new skills on success. Retrieval-augmented generation~\cite{lewis2020rag} retrieves raw past examples by embedding similarity. ISM differs along three axes: the bank stores compact symbolic schemas rather than executable code or raw examples, retrieval features and bank contents evolve through explicit self-maintenance mechanisms rather than monotonic growth, and every update is gated by symbolic verification before being committed.
\paragraph{Self-reflection and self-improvement.}
Reflection-based methods generate verbal critiques after failures. Reflexion~\cite{shinn2023reflexion} stores per-episode reflections in natural language, and Self-Refine~\cite{madaan2023selfrefineiterativerefinementselffeedback} iterates critique-and-revise within a single problem. STaR~\cite{zelikman2022star} bootstraps reasoning by fine-tuning on self-generated rationales. Most prior systems are asymmetric in their learning signal, accumulating either positive examples (STaR, Voyager) or negative feedback (Reflexion) but not both as structured, retrievable knowledge. ISM treats both as complementary signals through symmetric mechanisms: Self-Reinforce distills positive heuristics from successful episodes, while Self-Antipattern records mistakes to avoid from failures, with both injected into the solver prompt.
\paragraph{Mathematical reasoning and verification.}
Chain-of-thought prompting~\cite{wei2022cot} and program-aided variants~\cite{chen2022pot, gao2023pal} elicit step-by-step reasoning from frozen LLMs. Process reward models~\cite{lightman2023verify} score reasoning at the step level for verifier-guided generation. The MATH benchmark~\cite{hendrycks2021math} and competition-level evaluations such as OlympiadBench~\cite{he2024olympiadbench} have driven progress in this space. ISM combines schema-guided prompting with formal symbolic verification of intermediate steps and final answers, using verification not only on outputs but as a quality gate on what enters memory.

\section{Approach}

ISM consists of five components that interact during each episode: (i) a \textbf{feature extractor} that converts a problem into a structured representation, (ii) a \textbf{schema bank} where each schema is split into content and a feature hook, (iii) a \textbf{two-stage retrieval} procedure that selects the most relevant schema for the current problem, (iv) a \textbf{frozen LLM solver} that is prompted with the retrieved schema and may optionally invoke verifiable tools, and (v) a \textbf{memory controller} that runs a self-improvement loop over the bank after each episode. Figure~\ref{fig:pipeline} shows how these components interact. The remainder of this section formalizes the setting, describes each component, and details the verification-gated updates that distinguish ISM from passive memory baselines.

\subsection{Problem Setup}

We study continual mathematical reasoning under a streaming episode protocol. A problem stream $\mathcal{S} = \{(x_1, y_1), \dots, (x_T, y_T)\}$ is partitioned into $B$ blocks of $N$ problems each, where each block is drawn from a single mathematical domain (e.g., Algebra, Geometry, Number Theory). At episode $t$, the system observes problem $x_t$ and produces a candidate answer $\hat{y}_t$. After every block, the domain shifts.

Crucially, each episode is processed under \emph{hard episodic resets}: the model receives no context from previous episodes other than what is explicitly stored in the external memory $\mathcal{M}_t$. Formally, the solver computes
\begin{equation}
\hat{y}_t = f_\theta(x_t, \mathcal{M}_t),
\end{equation}
where $f_\theta$ is a frozen LLM with fixed parameters $\theta$, and $\mathcal{M}_t$ is the schema bank at the start of episode $t$. The bank is updated to $\mathcal{M}_{t+1}$ by the memory controller after observing the outcome. The frozen parameter constraint isolates the contribution of the external memory from any in-model adaptation.

We evaluate continual learning behavior with four standard metrics: average accuracy, plasticity (accuracy on the current block's domain), stability (accuracy on previously seen domains), forgetting (drop in accuracy on past domains relative to when they were last seen), and backward transfer (BWT, change in accuracy on past domains as new domains are added).

\subsection{Feature Extractor}
Each problem $x_t$ is mapped to a structured \textsc{ProblemFeatures} object that captures four attributes: an \emph{operator type} (algebraic, number theory, geometric, combinatoric, probability, or calculus), a \emph{structural pattern} (e.g., \texttt{two\_agents\_combined}, \texttt{optimization}, \texttt{evaluate\_expression}), a set of \emph{heuristic signatures} (e.g., \texttt{decompose}, \texttt{work\_backwards}, \texttt{apply\_theorems}), and a \emph{semantic embedding} of the problem text.

ISM uses a hybrid extractor that combines a rule-based branch (keyword matchers and surface-form patterns) with an LLM-based branch (a lightweight classification call). The two outputs are reconciled by an agreement score computed as a weighted sum over four attribute-level comparisons:
\begin{equation}
\begin{aligned}
\mathrm{conf} =\;& 0.40\,\mathbf{1}[o_r {=} o_l] + 0.30\,\mathbf{1}[p_r \in P_l^{\text{top-2}}] \\
&+ 0.15\,J(H_r, H_l) + 0.15\,\mathrm{sim}_q(q_r, q_l),
\end{aligned}
\label{eq:agreement}
\end{equation}
where subscripts $r$ and $l$ denote the rule-based and LLM-based extractions respectively, $o$ is the operator type, $p$ the structural pattern, $H$ the heuristic set, and $q$ the quantity signature. $J$ denotes Jaccard similarity over heuristic sets, and $\mathrm{sim}_q$ scores the quantity signature on attribute-level overlap (counts of known and unknown quantities, presence of rates, times, and constraints). The structural match is permissive: $p_r$ counts as matching if it appears in the LLM's top-2 predicted structures $P_l^{\text{top-2}}$. The four weights sum to one, so the threshold $\mathrm{conf} \geq 0.60$ requires the LLM extraction to match at least 60\% of the weighted attribute mass with the rule extractor. When this threshold is met, the LLM features are used; otherwise the rule-based features serve as a fallback. The output is a \texttt{ProblemFeatures} tuple $(o_t, p_t, H_t, q_t, e_t)$, where the embedding $e_t$ is computed independently by a sentence encoder, that is then passed to the schema bank.

\subsection{Two-Stage Retrieval}

Given problem features $F_t$, retrieval selects a schema in two stages.

\paragraph{Stage 1: Operator Filter.}
A hard filter restricts candidates to schemas whose hook \texttt{operator\_type} matches $F_t.\text{operator}$. This eliminates entire families of irrelevant schemas (e.g., geometry schemas for a number-theory problem) before scoring.

\paragraph{Stage 2: Soft Scoring.}
Each surviving candidate $s$ receives a weighted score combining structural, heuristic, quantity, embedding, and prior terms:
\begin{equation}
\text{score}(s, F_t) = \sum_{k \in K} w_k \cdot \text{sim}_k(s, F_t),
\label{eq:score}
\end{equation}
where $K = \{\text{structural}, \text{heuristic}, \text{quantity}, \text{embedding}, \text{prior}\}$ with weights $w = (0.15, 0.15, 0.05, 0.55, 0.10)$. The embedding similarity dominates because it generalizes across surface forms; structural and heuristic terms add complementary signal. The prior term incorporates the schema's success rate, encouraging the system to retrieve schemas with a track record.

\begin{figure*}[!t]
    \centering
    \includegraphics[width=1\linewidth]{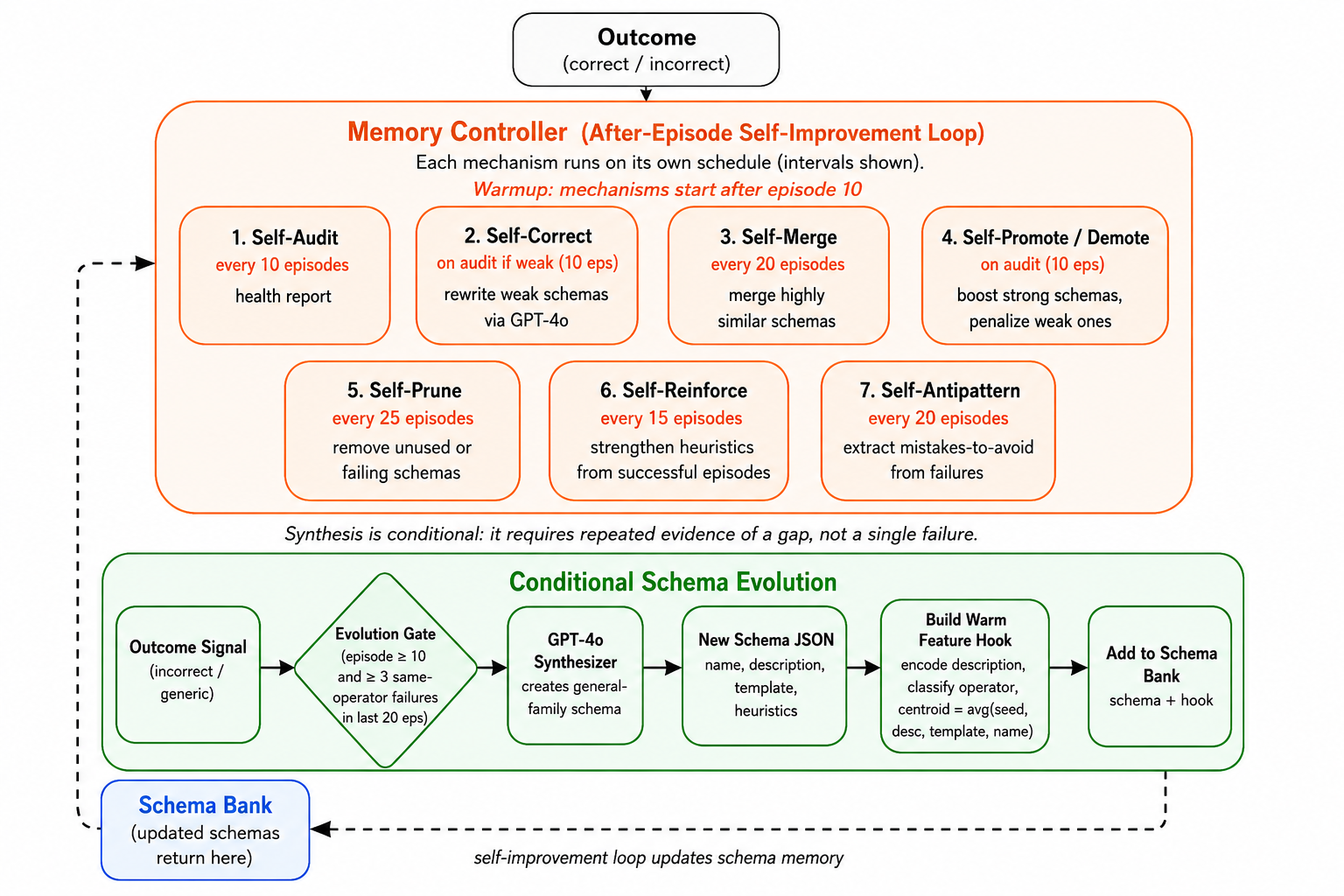}
    \caption{Memory Controller and conditional schema evolution. After each episode, the outcome is passed to the Memory Controller, which runs seven self-improvement mechanisms on independent schedules (top, orange): Self-Audit produces a health report every 10 episodes; Self-Correct rewrites weak schemas flagged by the audit; Self-Merge consolidates near-duplicates every 20 episodes; Self-Promote/Demote reweights schemas based on success rate; Self-Prune removes unused or failing schemas every 25 episodes; Self-Reinforce strengthens heuristics from successful episodes every 15; and Self-Antipattern extracts mistakes-to-avoid from failures every 20. Schema synthesis (bottom, green) is conditional: the Evolution Gate fires only when at least 10 episodes have elapsed and at least three same-operator failures have occurred in the last 20, after which a synthesizer creates a new general-family schema with a warm-started feature hook. The dashed arrow closes the self-improvement loop back to the Schema Bank.}
    \label{fig:memory_controller}
\end{figure*}

Based on the top score, the retrieval falls into one of three buckets: \emph{high} (score $\geq 0.80$, used directly), \emph{medium} (score $\geq 0.45$, used with a fallback generic schema attached), or \emph{generic} (score $< 0.45$, a domain-level generic schema is used and the outcome may trigger Conditional Schema Evolution).

\subsection{Frozen LLM Solver and Tools}

The retrieved schema's content is injected into a prompt template and passed to a frozen LLM solver via a \texttt{hard\_reset\_call}: a fresh call with no memory of previous episodes. The solver receives the problem text, the schema's description, solution template, and heuristics, and produces an answer.

For challenging problems, the solver can optionally enter an \emph{agentic mode} in which it invokes verifiable tools: a symbolic computation tool that evaluates expressions, a formal answer verifier built on symbolic algebra, a schema lookup tool that surfaces additional candidates, and a failure-search tool that retrieves antipatterns from the replay buffer. Tool use is bounded by a maximum-turn budget and only fires when the initial answer fails an internal sanity check. The agentic mode is gated rather than always-on, keeping inference cost low on routine problems.

\subsection{Memory Controller and Self-Improvement Loop}

After each episode, the memory controller performs two parallel tracks of updates.

\paragraph{Per-episode updates.}
The hook of the retrieved schema is updated after each episode via outcome-dependent exponential moving averages on the embedding centroid and success rate (full equations in Appendix~\ref{sec:schema_bank}).

\paragraph{Periodic self-improvement mechanisms.}
After a warmup of 10 episodes, seven mechanisms run on independent schedules to maintain the bank:

\begin{itemize}
    \item \textbf{Self-Audit} (every 10 eps): scores each schema as \emph{strong}, \emph{neutral}, \emph{weak}, or \emph{unused} based on success rate and usage count.
    \item \textbf{Self-Correct} (triggered by audit): when a schema is weak with at least 5 uses, an LLM rewrites its content and the success rate is reset to 0.5. After three failed corrections, the schema is escalated to pruning.
    \item \textbf{Self-Merge} (every 20 eps): pairs of schemas whose embedding centroids satisfy $\cos(\mu_i, \mu_j) > 0.88$ are merged via a weighted-centroid combination, with seed schemas protected.
    \item \textbf{Self-Promote / Demote} (triggered by audit): schemas with $\rho_s \geq 0.80$ have their retrieval scores boosted by 2\%; schemas with $\rho_s \leq 0.40$ are penalized by 5\%.
    \item \textbf{Self-Prune} (every 25 eps): non-seed schemas with zero uses or persistent failure are removed from the bank.
    \item \textbf{Self-Reinforce} (every 15 eps): for strong schemas, an LLM distills new positive heuristics from recent successful episodes and appends them to the schema content.
    \item \textbf{Self-Antipattern} (every 20 eps): for schemas associated with at least three failures, an LLM extracts mistakes-to-avoid from the failed episodes and records them in the content.
\end{itemize}

\paragraph{Conditional schema evolution.}
On incorrect or generic outcomes, the system evaluates an \emph{evolution gate}: if the current episode is at least 10 and at least three same-operator failures have occurred in the last 20 episodes, the synthesizer is invoked. The synthesizer is an LLM call that produces a new general-family schema in structured JSON form, after which a warm feature hook is built (operator classified, centroid initialized as the average of seed, description, template, and name embeddings) and the schema is added to the bank.

\noindent\textbf{Verification gate.}
Every memory update, including hook adjustment, merge, content rewrite, promote/demote, reinforce, antipattern, and synthesis, is gated by a symbolic verification step. For example, before Self-Reinforce appends new heuristics, the underlying solution must have been verified by symbolic computation, and before Self-Merge consolidates two schemas, their shared use cases must symbolically check out. This gate prevents the bank from reinforcing incorrect generalizations, which is the failure mode of unverified memory systems. For more additional details about the Schema Bank refer to Appendix~\ref{sec:schema_bank}

\section{Experiments}

\subsection{Benchmarks}
We evaluate on two competition-level mathematical reasoning benchmarks of different difficulty. \textbf{MATH-Hard} consists of the level-5 (hardest) problems from the MATH dataset~\cite{hendrycks2021math}, drawn from Algebra, Geometry, and Number Theory. These are high-school competition problems with short numerical or symbolic answers, hard enough to make a frozen LLM stumble often without good prompting. \textbf{OlympiadBench}~\cite{he2024olympiadbench} is a collection of olympiad-level mathematics problems, taken from international and national competitions, that are noticeably harder than MATH-Hard: longer setups, multi-step reasoning, and answers that often require careful case analysis or symbolic manipulation. For both benchmarks we run a 300-episode continual learning stream split into blocks by domain, with hard shifts between blocks. Every episode is processed under a hard episodic reset, so nothing carries over from one problem to the next except what the system has explicitly stored in its memory bank.

\subsection{Baselines}
We compare ISM against five baselines that cover the main approach families: (1)~\textbf{Vanilla LLM}, the frozen model with no external memory; (2)~\textbf{Static Schema}, one fixed prompt template used for every problem, with no retrieval and no updates; (3)~\textbf{RAG-over-Examples}, where every past episode is kept as an exemplar and the closest one by embedding similarity is fetched at inference time; (4)~\textbf{Reflexion}~\cite{shinn2023reflexion}, which adds a short verbal reflection to a growing buffer after each episode; and (5)~\textbf{Passive Schema Memory}, where schemas are created and stored just like in ISM but the seven self-improvement mechanisms are turned off. The gap between Passive and ISM tells us what active maintenance actually buys.
\subsection{Metrics}
We report final accuracy averaged over all 300 episodes along with four continual learning metrics. \emph{Plasticity} is the accuracy on the first 10 episodes of each new domain block, showing how quickly the system picks up a new domain. \emph{Stability} is the accuracy on the last 10 episodes of each domain, showing how well past performance is held on to. \emph{Forgetting} is how much accuracy drops on a domain compared to when it was last seen. \emph{Backward transfer} (BWT) tracks whether past domains get better or worse as new ones come in; a positive number means past domains keep improving. We also report the final bank size as a simple measure of storage cost.

\subsection{Implementation}

We use \texttt{gpt-4.1-mini} as the frozen solver across all systems, called via a hard-reset API call with no message history and temperature $0.0$. Self-improvement mechanisms that require an LLM (Self-Correct, Self-Merge, Self-Reinforce, Self-Antipattern, and schema synthesis) use \texttt{gpt-4o}, and problem embeddings come from \texttt{text-embedding-3-small}.Retrieval works in two stages. Stage~1 hard-filters schemas by operator type. Stage~2 ranks the remaining candidates by a weighted score over structural, heuristic, quantity, embedding, and prior terms with weights $(0.15, 0.15, 0.05, 0.55, 0.10)$. Retrievals fall into three buckets by top score: high ($\geq 0.80$, used directly), medium ($\geq 0.45$, used with a generic fallback), and generic ($< 0.45$).The hook of the retrieved schema is updated after each episode with outcome-dependent EMAs. The embedding centroid moves with $\alpha_e = 0.04$ on a success and $0.01$ on a failure, and the success rate moves with $\alpha_r = 0.15$ on a success and $0.10$ on a failure.After a warmup of 10 episodes, the seven self-improvement mechanisms run on independent schedules: Self-Audit every 10 episodes (which can trigger Self-Correct and Self-Promote/Demote), Self-Reinforce every 15, Self-Merge and Self-Antipattern every 20, and Self-Prune every 25. The evolution gate fires only when at least 10 episodes have elapsed and at least three same-operator failures have occurred in the last 20 episodes, so a new schema is synthesised only when the gap is real.

\subsection{Quantitative Results}
As seen in Table~\ref{tab:main}, the results on both benchmarks are brought together. ISM comes out ahead on accuracy, stability, forgetting, and backward transfer on both MATH-Hard and OlympiadBench, while keeping a much smaller bank than any retrieval baseline.

\begin{table*}[t]
\centering
\small
\caption{Main results on MATH-Hard and OlympiadBench over a 300-episode continual learning stream. Best values per benchmark in bold. ISM is best on accuracy, stability, and forgetting on both benchmarks, with positive backward transfer on OlympiadBench. RAG and Reflexion store one entry per episode, capped at the stream length.}
\label{tab:main}
\begin{tabular}{lcccccc}
\toprule
\textbf{System} & \textbf{Acc.} & \textbf{Plast.} & \textbf{Stab.} & \textbf{Forget.} & \textbf{BWT} & \textbf{Bank} \\
\midrule
\multicolumn{7}{l}{\emph{MATH-Hard}} \\
\midrule
Vanilla LLM & 48.00 & 46.00 & 50.00 & 0.10 & $-0.08$ & 0 \\
RAG-over-Examples & 57.00 & 54.00 & 60.00 & 0.12 & $-0.12$ & 300 \\
Reflexion & 55.67 & 52.67 & 58.67 & 0.13 & $-0.13$ & 300 \\
Static Schema & 78.67 & 76.00 & 81.33 & 0.08 & $-0.06$ & 1 \\
Passive Schema Memory & 78.67 & 75.33 & 82.00 & 0.10 & $-0.07$ & 47 \\
\textbf{ISM (ours)} & \textbf{80.67} & \textbf{76.67} & \textbf{84.67} & \textbf{0.07} & \textbf{$-0.05$} & \textbf{17} \\
\midrule
\multicolumn{7}{l}{\emph{OlympiadBench}} \\
\midrule
Vanilla LLM & 23.67 & 24.00 & 23.33 & 0.05 & $-0.03$ & 0 \\
Reflexion & 29.67 & 26.67 & 32.67 & 0.02 & $+0.01$ & 300 \\
RAG-over-Examples & 33.33 & 32.67 & 34.00 & 0.02 & $-0.02$ & 300 \\
Static Schema & 59.67 & 58.00 & 61.33 & 0.06 & $-0.00$ & 1 \\
Passive Schema Memory & 59.67 & 57.33 & 62.00 & 0.10 & $-0.02$ & 91 \\
\textbf{ISM (ours)} & \textbf{61.67} & \textbf{59.33} & \textbf{64.00} & \textbf{0.03} & \textbf{$+0.03$} & \textbf{13} \\
\bottomrule
\end{tabular}
\end{table*}

\paragraph{Accuracy and continual learning metrics.}
ISM reaches 80.67\% on MATH-Hard and 61.67\% on OlympiadBench, beating the strongest baseline (Passive Schema Memory) by exactly 2.00 points on each. The cumulative accuracy curves in Figures~\ref{fig:cumulative_math} and~\ref{fig:cumulative_olympiad} show ISM pulling ahead in the early episodes and holding the lead through every domain shift. Stability goes from 82.00\% to 84.67\% on MATH-Hard and from 62.00\% to 64.00\% on OlympiadBench, while forgetting drops from 0.10 to 0.07 and from 0.10 to 0.03 respectively, both compared to Passive. On OlympiadBench, ISM hits backward transfer of $+0.03$, meaning past domains actually keep getting better as new ones arrive. Reflexion is the only other system with a positive BWT on OlympiadBench ($+0.01$), but it sits at much lower accuracy overall.

\paragraph{Memory efficiency.}
ISM keeps 17 schemas to Passive's 47 on MATH-Hard (64\% fewer) and 13 to Passive's 91 on OlympiadBench (86\% fewer). The retrieval baselines hit 300 entries on both benchmarks. So ISM uses 17.6$\times$ fewer entries than RAG and Reflexion on MATH-Hard and 23$\times$ fewer on OlympiadBench. The gap grows with difficulty: harder problems push Passive to synthesize more schemas, but the merge and prune mechanisms in ISM keep the bank compact either way.

Figures~\ref{fig:bank_math} and~\ref{fig:bank_olympiad} show how the bank changes over time. On MATH-Hard, ISM's bank climbs during the early Geometry block (episodes 100--145) and then \emph{shrinks} from 26 down to 15 as Self-Prune and Self-Merge clean it up. Passive grows steadily from 6 to 47 with nothing pulling it back. On OlympiadBench, the gap is even bigger: ISM stays in the 10--22 range the whole stream, while Passive climbs linearly to 91. The repeated dips in the ISM curve are the prune and merge cycles doing their job; without them, the bank would look like the Passive curve.

\subsection{Qualitative Results}
To see where active memory actually helps, we picked out episodes where ISM got the right answer while three or more baselines got it wrong.

Each of these schemas was synthesized after the evolution gate caught a repeated pattern of same-operator failures. After that, they were reused 5, 12, and 19 times respectively, with success rates between 0.63 and 0.76. When the same kind of pattern shows up again later in the stream, ISM already has something ready to apply. On Episode 235 the Passive baseline returned 121 (wrong) while ISM returned 93 (right): both had a relative-primality schema in their bank, but only ISM had spent the episodes in between sharpening its feature hook through self-improvement to actually retrieve it when it mattered.For additional quantitative results and qualitative case studies, refer to Appendix~\ref{sec:case_studies}.

\section{Limitations}

This work presents an initial study of Intelligent Schema Memory (ISM) under continual mathematical reasoning with frozen LLMs, and several limitations remain.

First, all experiments are conducted using a single random seed and a single ordering of the continual learning stream. Continual learning systems are known to be sensitive to both initialization and stream composition, since different domain orders alter which transitions stress adaptation and forgetting. As a result, the reported gains should be interpreted as preliminary evidence rather than definitive estimates of performance. Multi-seed evaluation, shuffled stream orderings, and variance reporting are important next steps that we leave for future work.

Second, we do not include per-mechanism ablations isolating the contribution of individual components such as Self-Reinforce, Self-Antipattern, Self-Merge, Self-Prune, the verification gate, or conditional schema synthesis. Instead, we evaluate an aggregate ablation through the Passive Schema Memory baseline, where all self-improvement mechanisms are disabled simultaneously. While this comparison demonstrates the overall value of active memory maintenance, it does not clarify which mechanisms contribute most strongly to the observed improvements or which may be redundant.

Third, our baselines focus primarily on memory-oriented approaches, including retrieval-based storage, reflection buffers, and passive schema memory. We do not compare against search-heavy reasoning systems such as Tree-of-Thoughts \cite{yao2023tree}, graph-structured memory architectures \cite{oliveira2025knowledge}, or fully agentic ReAct-style frameworks \cite{yao2022react}. These systems target somewhat different aspects of reasoning and control, and a broader comparison across reasoning paradigms would strengthen the evaluation.

In addition, all experiments are performed using a single frozen LLM backbone. Although ISM is designed to be model-agnostic, we have not yet evaluated whether the same trends hold consistently across smaller or larger models. The evaluation is also limited to two competition-level mathematical reasoning benchmarks. Whether the same lifecycle dynamics and symmetric success/failure learning mechanisms generalize to other domains, such as formal proof generation, code reasoning, or broader scientific problem solving, remains an open question.

Finally, the verification gate currently relies on symbolic computation tools that are well suited for numerical and short symbolic answers, but less applicable to proof-oriented tasks where correctness itself may require complex reasoning. Extending verification to richer forms of mathematical reasoning is an important direction for future work. Similarly, the agentic tool-use loop in ISM is selectively activated only after failed initial attempts, and we do not evaluate settings where tool use is persistent or always-on.
\section{Conclusion}

We presented \textbf{Intelligent Schema Memory (ISM)}, a self-evolving memory system that lets a frozen LLM improve at mathematical reasoning under hard episodic resets. ISM keeps a compact bank of strategy schemas, each split into content that goes into the solver prompt and a feature hook that adapts online, with a self-improvement loop that audits, merges, prunes, reinforces, and records antipatterns, all gated by symbolic verification. Across MATH-Hard and OlympiadBench, ISM achieves the strongest accuracy, stability, and forgetting metrics among the systems we tested, while keeping a memory bank 64 to 86 percent smaller than the strongest non-active baseline and up to 23 times smaller than retrieval-based systems.

These results suggest that small, actively maintained, verifiable strategy memory is a viable path for continual reasoning in frozen-LLM systems. The next steps are multi-seed evaluation, per-mechanism ablations, and extending ISM beyond mathematical reasoning to domains where verification is less direct.

\bibliography{example_paper}
\bibliographystyle{abbrvnat}
\clearpage
\appendix
\onecolumn

\section{Appendix}

\subsection{Schema Bank}
\label{sec:schema_bank}

The schema bank $\mathcal{M}_t$ stores a set of \emph{strategy schemas}. Each schema $s$ is split into two complementary parts:

\begin{itemize}
    \item \textbf{Content section}, containing the schema's name, natural-language description, solution template, and a set of heuristics. The content is what gets injected into the solver's prompt and is largely stable over time.
    \item \textbf{Feature hook}, containing the retrieval-side signals: \textit{operator type}, \textit{structural pattern}, \textit{heuristic signature}, \textit{quantity signature}, an \textit{embedding centroid} (maintained as an exponential moving average over the embeddings of problems the schema has handled), a \textit{success rate} (also EMA), and \textit{usage count}.
\end{itemize}

The hook adapts online while the content stays fixed unless rewritten by Self-Correct. This separation lets retrieval sharpen with experience without destabilizing the strategies themselves. After each use of schema $s$ on problem $x_t$ with embedding $e_t$ and outcome $r_t \in \{0, 1\}$, the hook updates its centroid and success rate via outcome-dependent EMAs:
\begin{align}
\mu_s &\leftarrow (1 - \alpha_e^{(r_t)}) \mu_s + \alpha_e^{(r_t)} e_t, \\
\rho_s &\leftarrow (1 - \alpha_r^{(r_t)}) \rho_s + \alpha_r^{(r_t)} r_t,
\end{align}
where the rates are deliberately asymmetric. The centroid uses $\alpha_e^{(\text{correct})}=0.04$ and $\alpha_e^{(\text{wrong})}=0.01$: successful retrievals pull the centroid harder toward exemplars the schema handled well, while failed retrievals leave the centroid largely intact. The success rate uses $\alpha_r^{(\text{correct})}=0.15$ and $\alpha_r^{(\text{wrong})}=0.10$: positive evidence registers slightly faster than negative evidence, so newly-strong schemas are recognized quickly while occasional misses do not over-penalize them.

\begin{figure*}[t]
\centering
\includegraphics[width=\textwidth]{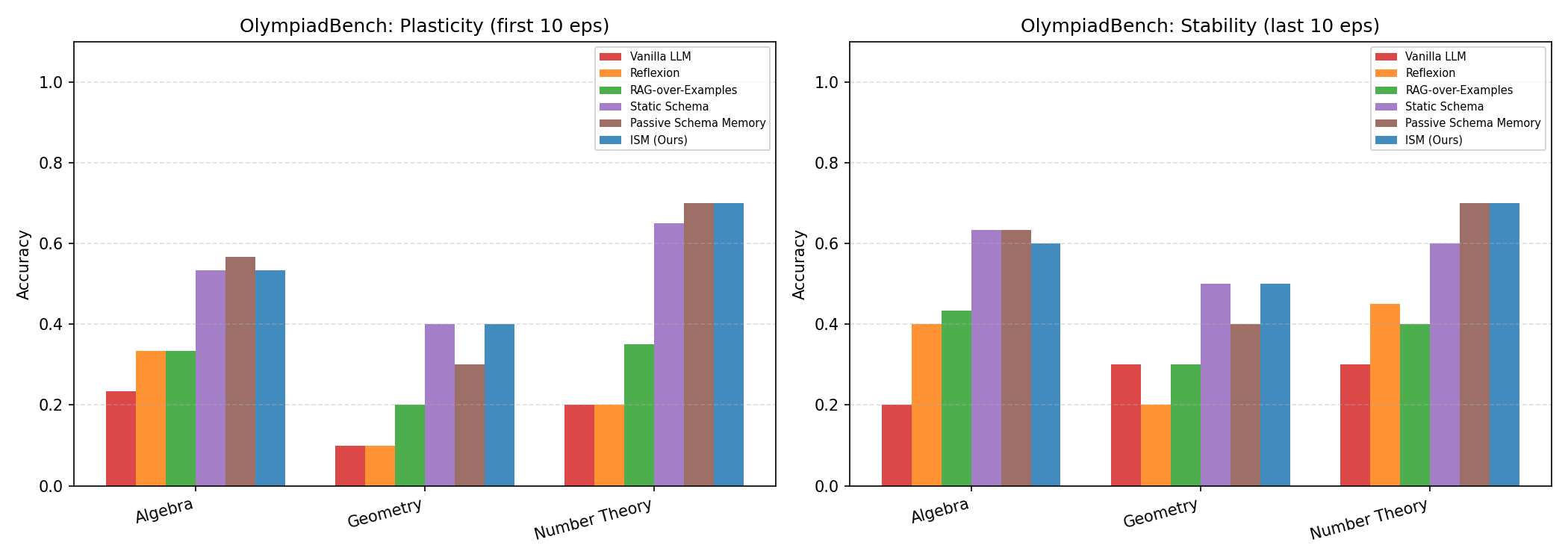}
\caption{OlympiadBench plasticity (first 10 episodes of each block, left) and stability (last 10 episodes, right) per domain. ISM matches or beats Passive Memory on both metrics across Algebra, Geometry, and Number Theory.}
\label{fig:plast_stab_olympiad}
\end{figure*}

\begin{figure*}[t]
\centering
\includegraphics[width=\textwidth]{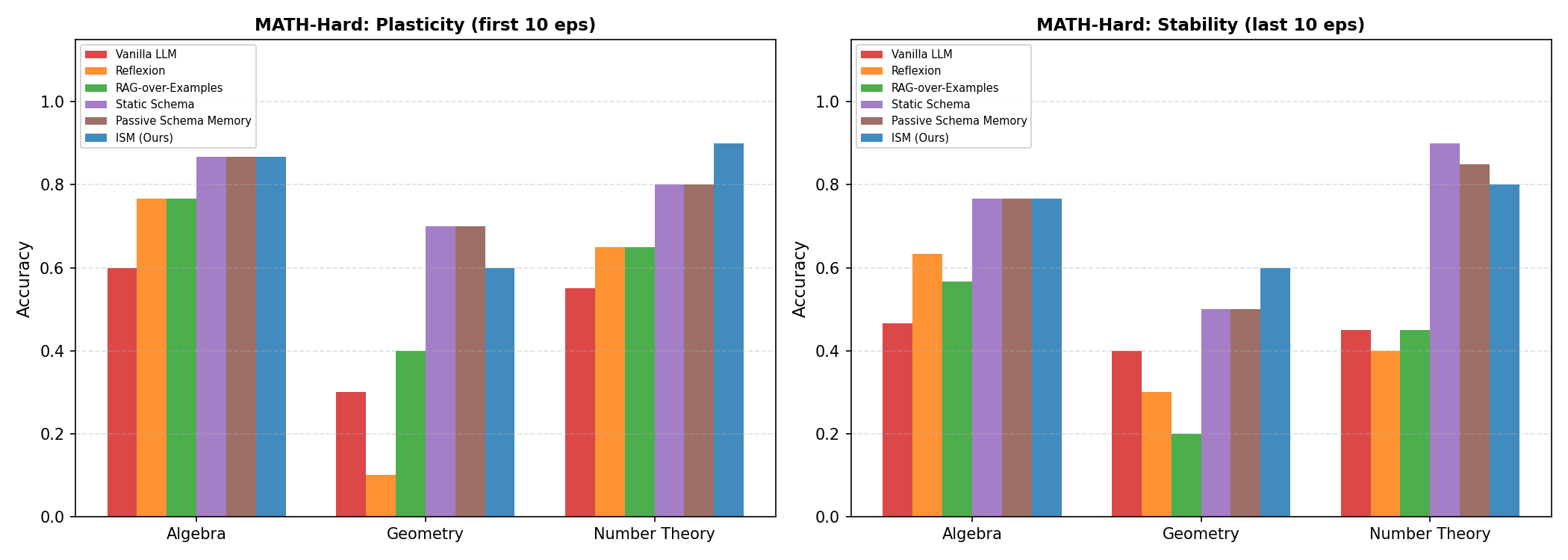}
\caption{MATH-Hard plasticity (first 10 episodes of each block, left) and stability (last 10 episodes, right) per domain.}
\label{fig:plast_stab_math}
\end{figure*}

\paragraph{Per-domain plasticity and stability.}
Figure~\ref{fig:plast_stab_olympiad} breaks plasticity and stability down by domain on OlympiadBench. ISM matches or beats Passive on plasticity in every domain, with the biggest gain in Geometry (40\% vs 30\%). On stability, ISM holds its own with the best baselines and pulls ahead in Number Theory. The same picture shows up on MATH-Hard (Figure~\ref{fig:plast_stab_math}): ISM ties Static and Passive on Algebra, wins on Geometry stability, and slips a bit on Number Theory stability.

\begin{figure*}[t]
\centering
\includegraphics[width=\textwidth]{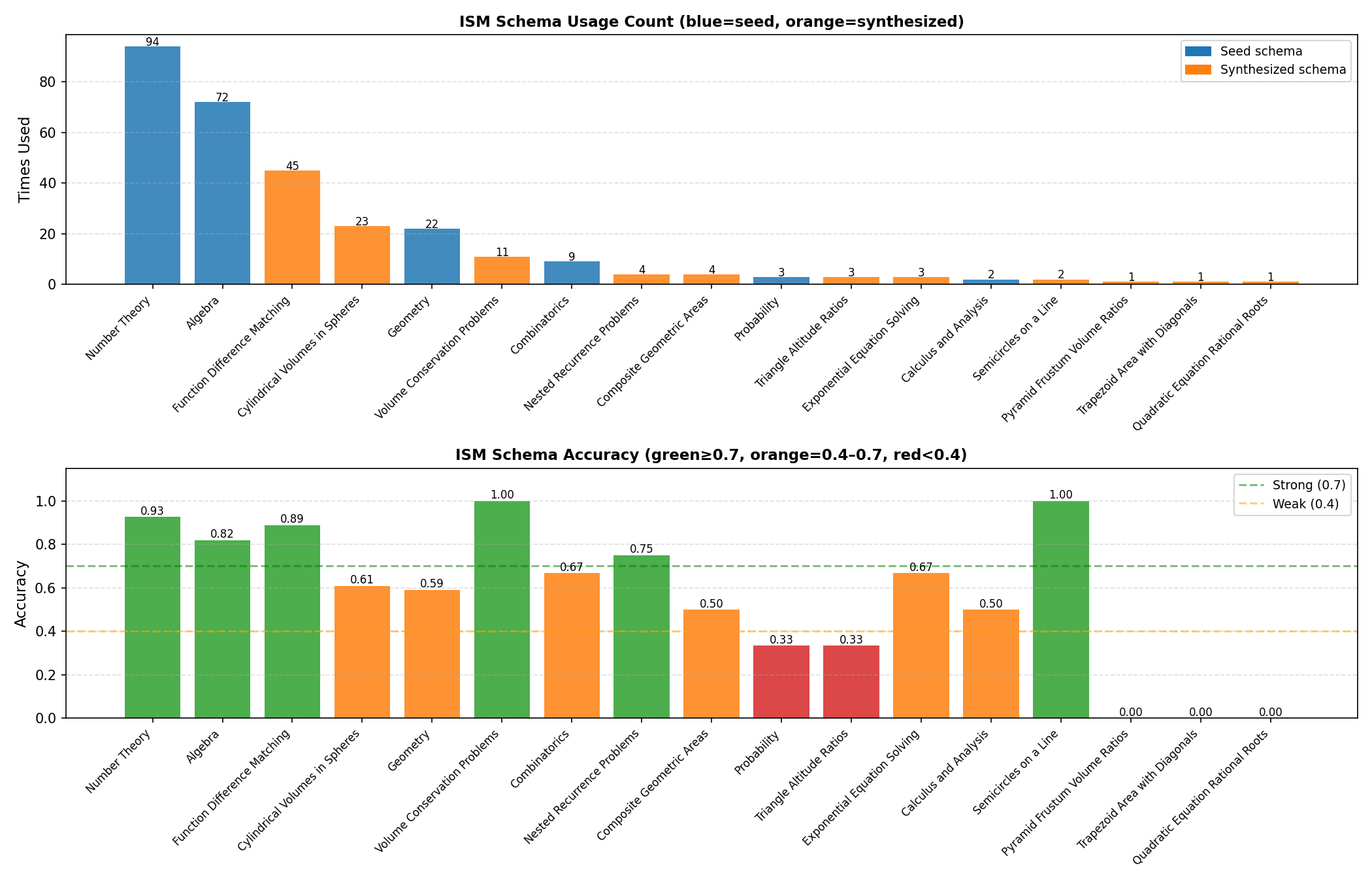}
\caption{ISM's schema bank on MATH-Hard at the end of the stream. Top: usage count per schema (blue = seed, orange = synthesized). Bottom: success rate per schema. Most retained schemas sit above the strong-schema threshold of 0.7 (green dashed line).}
\label{fig:schema_usage_math}
\end{figure*}

\begin{figure*}[t]
\centering
\includegraphics[width=\textwidth]{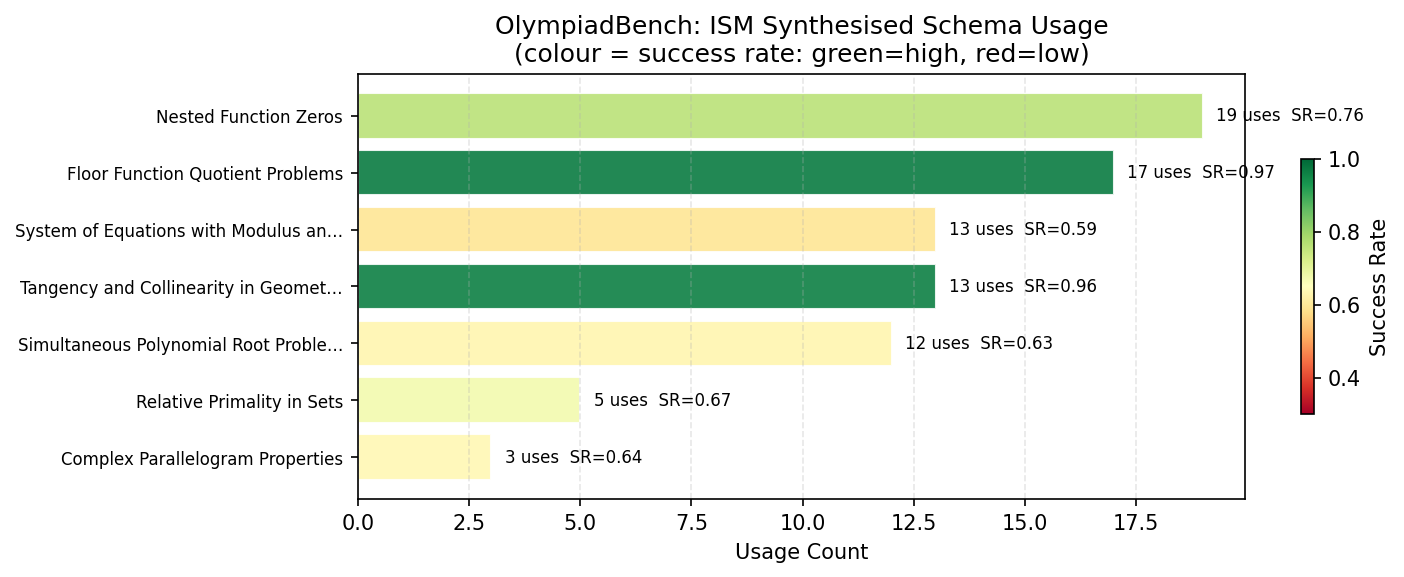}
\caption{ISM synthesized schemas on OlympiadBench. Every retained synthesized schema has at least three uses and a success rate above 0.59. Color shows success rate (green = high, red = low).}
\label{fig:schema_usage_olympiad}
\end{figure*}

\paragraph{Schema Bank Composition}
The composition of the two banks looks very different. On MATH-Hard, Passive ends with 47 schemas, but 28 of them (60\%) were never used again after being created. On OlympiadBench, Passive ends with 91 schemas and 62 of them (68\%) were never used again. These are schemas that were synthesized once on a failure and then never retrieved, so they take up space without giving anything back. Figure~\ref{fig:schema_usage_math} shows what ISM's bank looks like on MATH-Hard: the seed schemas (Number Theory, Algebra, Geometry, Combinatorics) carry most of the load, but synthesized schemas earn their place too, with \emph{Function Difference Matching} hitting 45 uses at 89\% accuracy. Figure~\ref{fig:schema_usage_olympiad} shows the same picture on OlympiadBench: every synthesized schema that survived has at least 3 uses and a success rate above 0.59, with \emph{Floor Function Quotient Problems} and \emph{Tangency and Collinearity} above 0.96. Self-Prune cuts off the long tail of unused schemas that Passive holds on to forever.

\section{Wins and losses against each baseline.}

\begin{figure}
    \centering
    \includegraphics[width=1\linewidth]{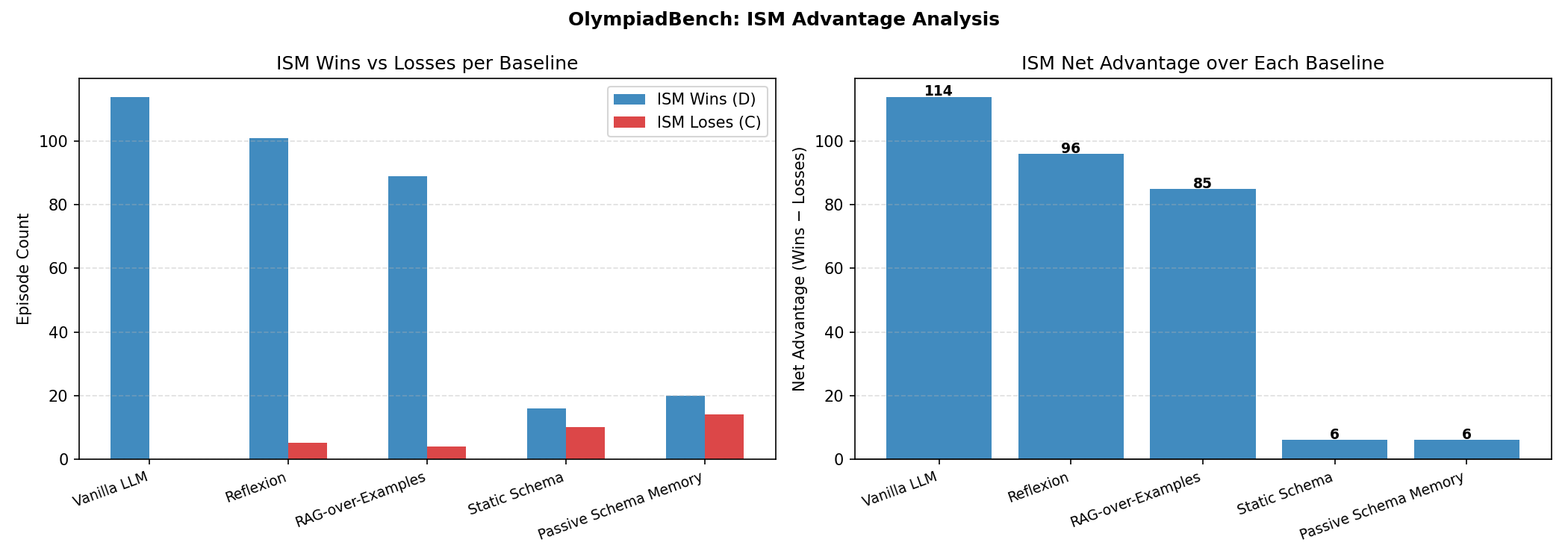}
    \caption{ISM head-to-head against each baseline on OlympiadBench. \textbf{Left:} wins (ISM correct, baseline wrong) versus losses (baseline correct, ISM wrong) per baseline. \textbf{Right:} net advantage (wins minus losses). ISM dominates the unmanaged-memory baselines ($+114$ over Vanilla, $+96$ over Reflexion, $+85$ over RAG) and holds a steady $+6$ lead over both schema-based controls (Static, Passive). The consistent $+6$ over the strongest baselines aligns with the 2-point accuracy gap}
    \label{fig:ism_olympiad}
\end{figure}

\begin{figure}
    \centering
    \includegraphics[width=1\linewidth]{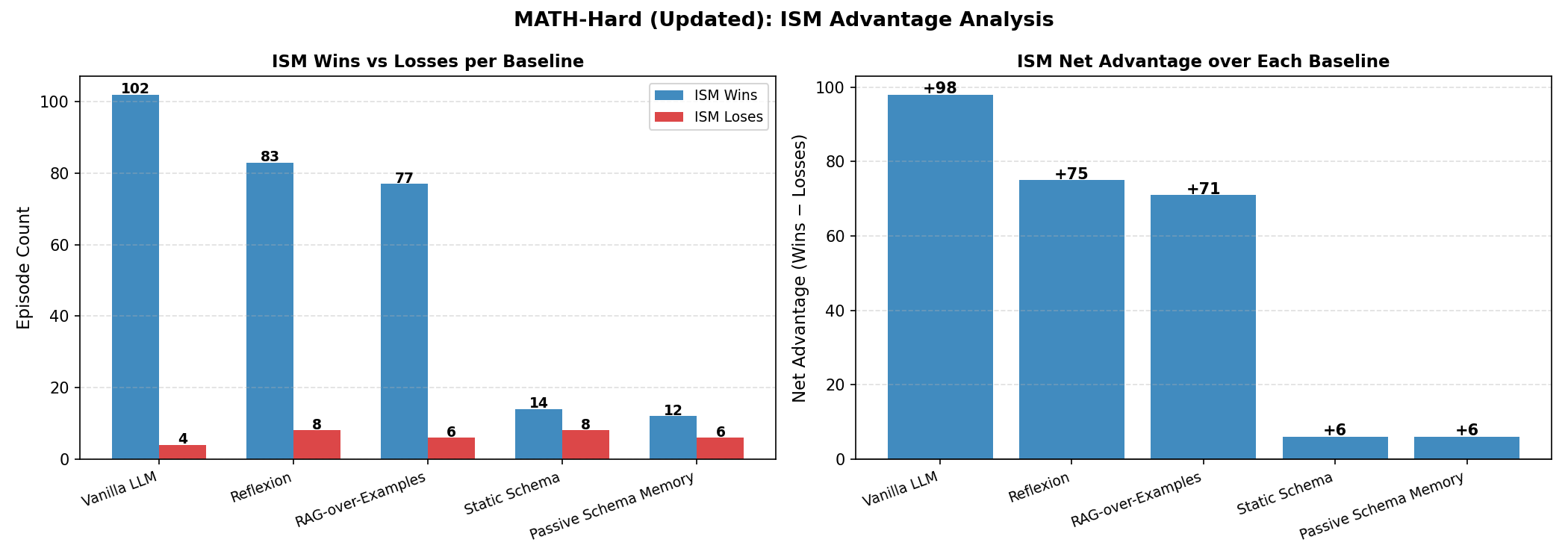}
    \caption{ISM head-to-head against each baseline on MATH-Hard. \textbf{Left:} wins versus losses per baseline. \textbf{Right:} net advantage. The pattern mirrors OlympiadBench: large net leads over the unmanaged baselines ($+98$ over Vanilla, $+75$ over Reflexion, $+71$ over RAG) and a $+6$ lead over both schema-based controls (Static, Passive).}
    \label{fig:ism_math}
\end{figure}
Figure~\ref{fig:ism_olympiad} compares ISM head-to-head with each baseline on OlympiadBench. Against the unmanaged-memory baselines the lead is huge: $+114$ episodes over Vanilla, $+96$ over Reflexion, $+85$ over RAG. Against the schema-based controls the lead is small but consistent: $+6$ over Static and $+6$ over Passive. The MATH-Hard picture (Figure~\ref{fig:ism_math}) is the same shape: $+98$, $+75$, $+71$ over the unmanaged baselines, and $+6$ each over Static and Passive. The steady $+6$ over the strongest controls is exactly the 2-point gap in the table.

\section{Qualitative Case Studies}
\label{sec:case_studies}

We present five representative episodes where ISM produced the correct answer through a \emph{synthesised} schema while several baselines failed. These cases illustrate how the evolution gate produces durable, transferable strategies rather than one-shot patches: every schema below was created mid-stream by the synthesiser, accumulated meaningful reuse, and reached a positive success rate.

\subsection{OlympiadBench Case Studies}

\subsubsection*{Episode 213 (Number Theory)}

\noindent\textbf{Problem.}
\begin{quote}
\itshape
Let $p$ be a prime number. If $p$ years ago, the ages of three children formed a geometric sequence with a sum of $p$ and a common ratio of 2, compute the sum of the children's current ages.
\end{quote}

\noindent\textbf{Schema retrieved.} \texttt{Simultaneous Polynomial Root Problems} (synthesised; success rate $0.63$, used 12 times).

\noindent\textbf{Schema description.} \emph{Determine relationships between the roots of multiple polynomial equations with constraints or shared roots.}

\noindent\textbf{Heuristics applied.} introduce variable; decompose; use symmetry.

\noindent\textbf{Outcome.} ISM produced the correct answer $\boxed{28}$. All four baselines answered incorrectly (e.g., Vanilla LLM predicted $4p$).

\subsubsection*{Episode 235 (Number Theory)}

\noindent\textbf{Problem.}
\begin{quote}
\itshape
Compute the third least positive integer $n$ such that each of $n$, $n+1$, and $n+2$ is a product of exactly two (not necessarily distinct) primes.
\end{quote}

\noindent\textbf{Schema retrieved.} \texttt{Relative Primality in Sets} (synthesised; success rate $0.67$, used 5 times).

\noindent\textbf{Schema description.} \emph{Find the smallest set of integers constructed under specific rules where no element is relatively prime to the product of the others.}

\noindent\textbf{Heuristics applied.} decompose; find pattern; introduce variable.

\noindent\textbf{Outcome.} ISM produced the correct answer $\boxed{93}$. All four baselines answered incorrectly (e.g., Passive Schema Memory predicted $121$).

\subsubsection*{Episode 286 (Algebra)}

\noindent\textbf{Problem.}
\begin{quote}
\itshape
The equations $x^{3}+Ax+10=0$ and $x^{3}+Bx^{2}+50=0$ have two roots in common. Compute the product of these common roots.
\end{quote}

\noindent\textbf{Schema retrieved.} \texttt{Nested Function Zeros} (synthesised; success rate $0.76$, used 19 times).

\noindent\textbf{Schema description.} \emph{Solve for the unknown polynomials and their compositions when dealing with nested polynomial equations and specific shared-root conditions.}

\noindent\textbf{Heuristics applied.} analyse degrees and symmetry; consider boundary or extreme values; use composition properties creatively.

\noindent\textbf{Outcome.} ISM produced the correct answer $\boxed{5\sqrt[3]{4}}$. All four baselines failed to produce a valid answer.

\subsection{MATH-Hard Case Studies}

\subsubsection*{Episode 122 (Geometry)}

\noindent\textbf{Problem.}
\begin{quote}
\itshape
In the diagram, two pairs of identical isosceles triangles are cut off of square $ABCD$, leaving rectangle $PQRS$. The total area cut off is $200~\text{m}^2$. What is the length of $PR$, in meters?
\end{quote}

\noindent\textbf{Schema retrieved.} \texttt{Composite Geometric Areas} (synthesised; success rate $0.61$, used 5 times).

\noindent\textbf{Schema description.} \emph{Problems involving the calculation of areas formed by overlapping geometric figures such as squares, circles, and triangles.}

\noindent\textbf{Heuristics applied.} decompose; change representation; use symmetry.

\noindent\textbf{Outcome.} ISM produced the correct answer $\boxed{20}$. All four baselines answered incorrectly (e.g., Passive Schema Memory predicted $5\sqrt{10}$).

\subsubsection*{Episode 154 (Algebra)}

\noindent\textbf{Problem.}
\begin{quote}
\itshape
Let $f$ be defined by
\[
f(x) = 
\begin{cases}
2 - x & \text{if } x \leq 1, \\
2x - x^2 & \text{if } x > 1.
\end{cases}
\]
Calculate $f^{-1}(-3) + f^{-1}(0) + f^{-1}(3)$.
\end{quote}

\noindent\textbf{Schema retrieved.} \texttt{Function Difference Matching} (synthesised; success rate $0.97$, used 85 times).

\noindent\textbf{Schema description.} \emph{Problems involving determining a parameter such that the difference between the evaluations of two functions at a target value matches a given constraint.}

\noindent\textbf{Heuristics applied.} decompose; introduce variable; work backwards.

\noindent\textbf{Outcome.} ISM produced the correct answer $\boxed{4}$. All three baselines failed to produce a valid answer.

\begin{figure*}[t]
\centering
\includegraphics[width=\textwidth]{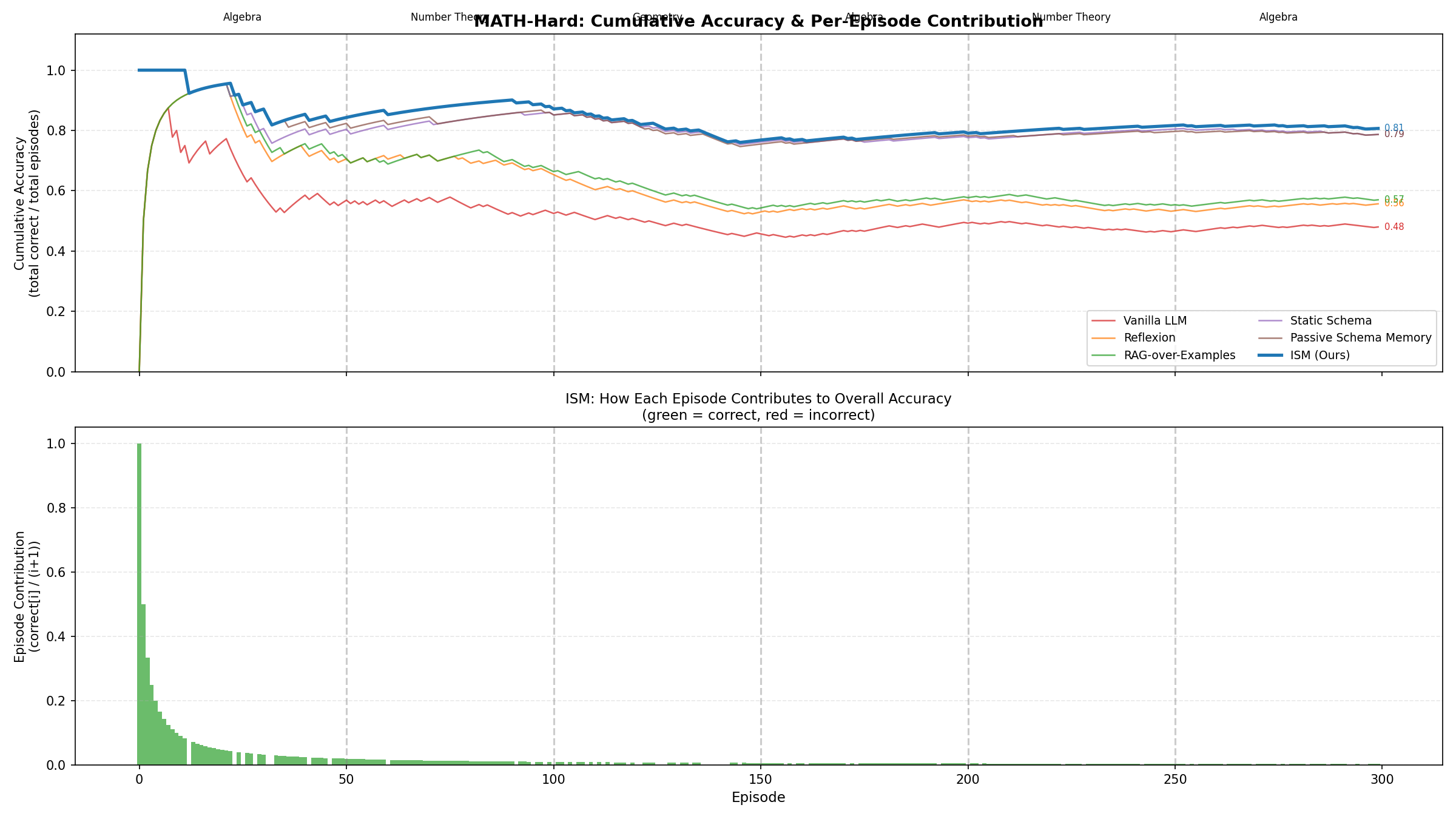}
\caption{Cumulative accuracy across the 300-episode MATH-Hard stream. ISM leads from early episodes onward and stays ahead through every domain transition (dashed vertical lines), finishing at 0.81. Static Schema and Passive Memory finish at 0.79; RAG and Reflexion stay near 0.56; Vanilla LLM ends at 0.48.}
\label{fig:cumulative_math}
\end{figure*}

\begin{figure*}[t]
\centering
\includegraphics[width=\textwidth]{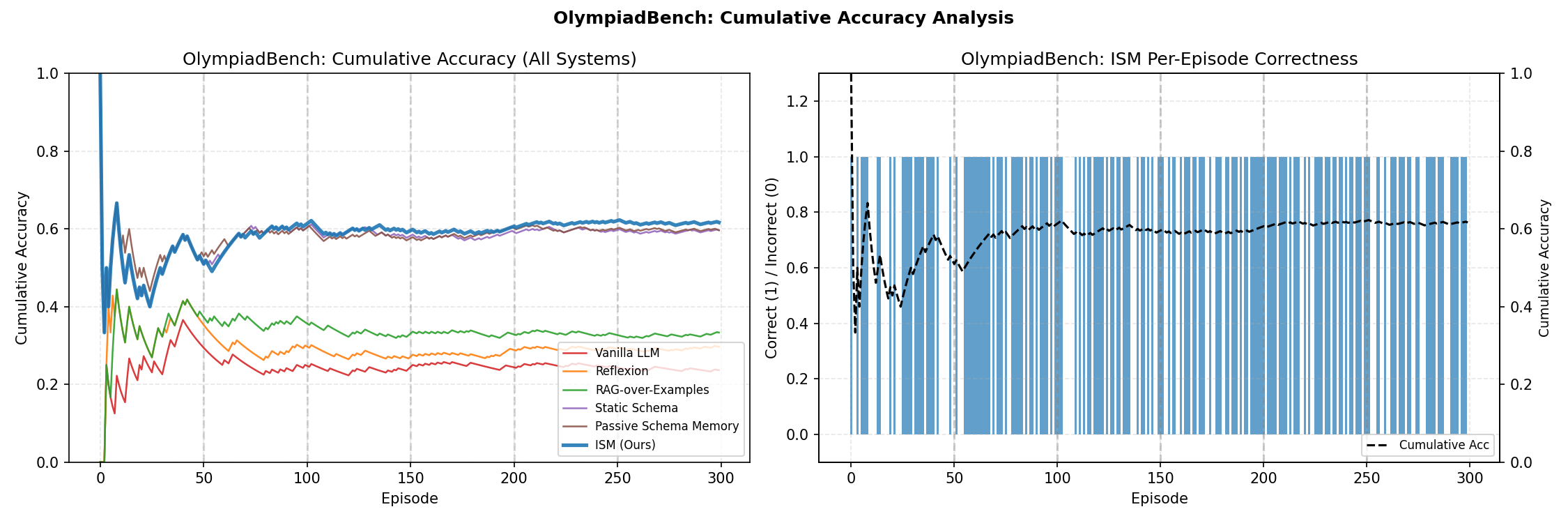}
\caption{Cumulative accuracy on OlympiadBench (left) and ISM's per-episode correctness with running cumulative accuracy (right). The schema-based systems (ISM, Passive, Static) cluster around 0.60, with ISM slightly above. RAG, Reflexion, and Vanilla stay below 0.35.}
\label{fig:cumulative_olympiad}
\end{figure*}

\begin{figure*}[t]
\centering
\includegraphics[width=\textwidth]{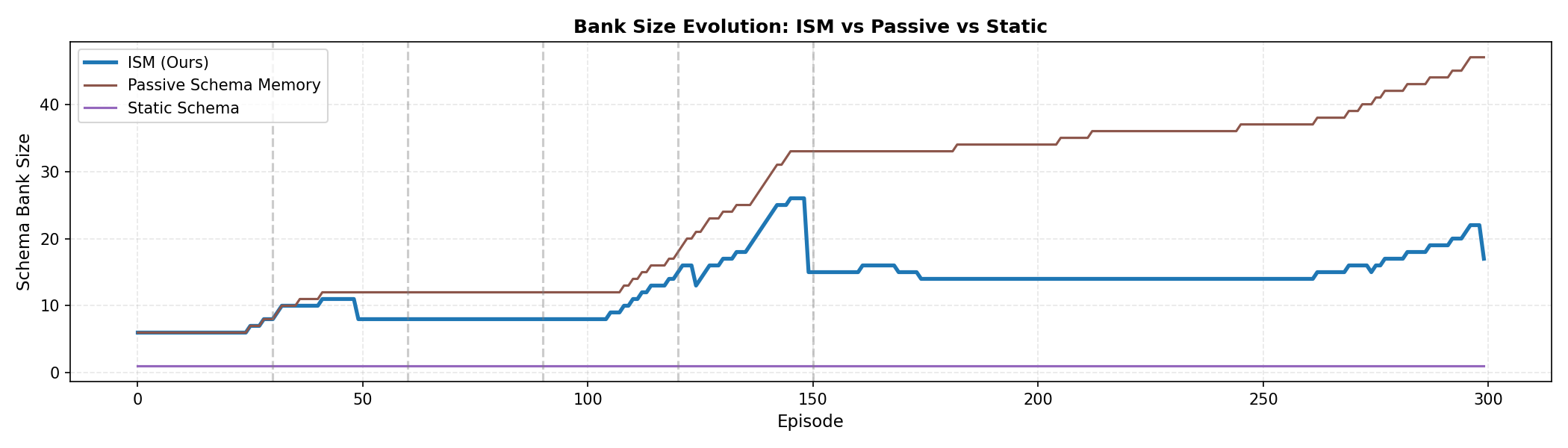}
\caption{Schema bank size over the MATH-Hard stream. ISM's bank grows during exploration and then drops when Self-Prune and Self-Merge fire (around episode 150 and again past 270). Passive keeps growing.}
\label{fig:bank_math}
\end{figure*}

\begin{figure*}[t]
\centering
\includegraphics[width=\textwidth]{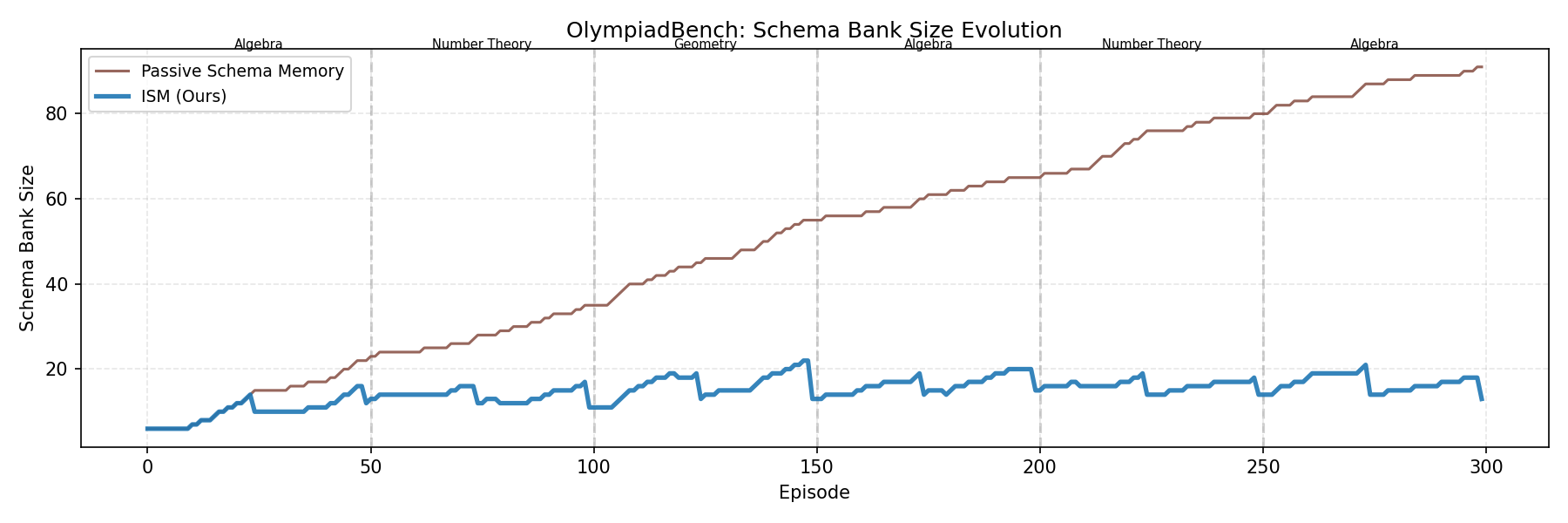}
\caption{Schema bank size over the OlympiadBench stream. With lifecycle management, the ISM bank stays in the 10--22 range, while Passive climbs to 91. Domain labels at top show which subject each block is drawn from.}
\label{fig:bank_olympiad}
\end{figure*}

\subsection{Hyperparameters}

\begin{table}[t]
\centering
\small
\renewcommand{\arraystretch}{1.12}
\setlength{\tabcolsep}{8pt}

\begin{tabular}{p{0.62\linewidth}c}
\toprule
\textbf{Parameter} & \textbf{Value} \\
\midrule

\multicolumn{2}{l}{\textit{Models}} \\[2pt]
Solver LLM & gpt-4.1-mini \\
Synthesizer LLM & gpt-4o \\
Embedding model & text-embedding-3-small \\

\addlinespace
\multicolumn{2}{l}{\textit{Feature extractor}} \\[2pt]
Agreement threshold & 0.60 \\

\addlinespace
\multicolumn{2}{l}{\textit{Retrieval scoring weights}} \\[2pt]
Operator filter weight & 0.00 (hard filter) \\
Structural weight & 0.15 \\
Heuristic weight & 0.15 \\
Quantity weight & 0.05 \\
Embedding weight & 0.55 \\
Prior (success rate) weight & 0.10 \\

\addlinespace
\multicolumn{2}{l}{\textit{Retrieval buckets}} \\[2pt]
High confidence & $\geq 0.80$ \\
Medium confidence & $\geq 0.45$ \\
Generic & $< 0.45$ \\

\addlinespace
\multicolumn{2}{l}{\textit{EMA rates (centroid)}} \\[2pt]
Success $\alpha_e$ & 0.04 \\
Failure $\alpha_e$ & 0.01 \\

\addlinespace
\multicolumn{2}{l}{\textit{EMA rates (success rate)}} \\[2pt]
Success $\alpha_r$ & 0.15 \\
Failure $\alpha_r$ & 0.10 \\

\addlinespace
\multicolumn{2}{l}{\textit{Mechanism intervals (episodes)}} \\[2pt]
Self-Audit & 10 \\
Self-Reinforce & 15 \\
Self-Merge & 20 \\
Self-Antipattern & 20 \\
Self-Prune & 25 \\
Replay check & 10 \\

\addlinespace
\multicolumn{2}{l}{\textit{Mechanism thresholds}} \\[2pt]
Warmup episodes & 10 \\
Self-Correct min uses & 5 \\
Self-Correct fail-then-prune & 3 \\
Self-Merge cosine threshold & 0.88 \\
Self-Promote success rate & $\geq 0.80$ \\
Self-Promote boost & $+2\%$ \\
Self-Demote success rate & $\leq 0.40$ \\
Self-Demote penalty & $-5\%$ \\

\addlinespace
\multicolumn{2}{l}{\textit{Evolution gate}} \\[2pt]
Minimum episode & 10 \\
Failures in window & $\geq 3$ in last 20 \\

\addlinespace
\multicolumn{2}{l}{\textit{Replay buffer}} \\[2pt]
Max size & 500 \\
Failure bias on sampling & 70\% \\

\bottomrule
\end{tabular}

\caption{Full hyperparameter configuration for ISM.}
\label{tab:ism_hyperparameters}

\end{table}
\end{document}